\documentclass[lettersize,journal]{IEEEtran}
\usepackage{amsmath,amsfonts}
\usepackage{algorithm}
\usepackage{array}
\usepackage[caption=false,font=normalsize,labelfont=sf,textfont=sf]{subfig}
\usepackage{textcomp}
\usepackage{stfloats}
\usepackage{url}
\usepackage{verbatim}
\usepackage{graphicx}
\usepackage{cite}

\usepackage{booktabs}
\usepackage{multirow}
\usepackage{tabularx}
\usepackage{bm}
\usepackage{hyperref}

\usepackage{algpseudocode}
\usepackage{amsmath}
\usepackage{color}
\usepackage{makecell}
\hypersetup{
colorlinks=true,
citecolor=black,
linkcolor=black
}

\begin{document}


\title{Insight Any Instance: Promptable Instance Segmentation for Remote Sensing Images}
\author{Xuexue Li, Wenhui Diao, Xinming Li and Xian Sun,~\IEEEmembership{Senior Member, IEEE}

\thanks{This work was supported by the National Key R\&D Program of China under 2022ZD0118402.
\textit{ (Corresponding author: Wenhui Diao.) }
}
\thanks{Xuexue Li, Wenhui Diao, Xinming Li and Xian Sun are with the Aerospace Informa-
tion Research Institute, Chinese Academy of Sciences, Beijing 100190, China,
and also with the Key Laboratory of Network Information System Technol-
ogy (NIST), Aerospace Information Research Institute, Chinese Academy
of Sciences, Beijing 100190, China, also with the University of Chinese
Academy of Sciences, Beijing 100190, China, and also with the School of
Electronic, Electrical and Communication Engineering, University of Chinese
Academy of Sciences, Beijing 100190, China (e-mail: lixuexue20@mails.ucas.ac.cn).}
}



\maketitle

\begin{abstract}
Instance segmentation of remote sensing images (RSIs) is an essential task for a wide range of applications such as land planning and intelligent transport.
Instance segmentation of RSIs is constantly plagued by the unbalanced ratio of foreground and background and limited instance size. And most of the instance segmentation models are based on deep feature learning and contain operations such as multiple downsampling, which is harmful to instance segmentation of RSIs, and thus the performance is still limited. 
Inspired by the recent superior performance of prompt learning in visual tasks, we propose a new prompt paradigm to address the above issues.
Based on the existing instance segmentation model, firstly, a local prompt module is designed to mine local prompt information from original local tokens for specific instances; secondly, a global-to-local prompt module is designed to model the contextual information from the global tokens to the local tokens where the instances are located for specific instances. Finally, a proposal's area loss function is designed to add a decoupling dimension for proposals on the scale to better exploit the potential of the above two prompt modules.
It is worth mentioning that our proposed approach can extend the instance segmentation model to a promptable instance segmentation model, i.e., to segment the instances with the specific boxes prompt. The time consumption for each promptable instance segmentation process is only 40 ms.
The paper evaluates the effectiveness of our proposed approach based on several existing models in four instance segmentation datasets of RSIs, and thorough experiments prove that our proposed approach is effective for addressing the above issues and is a competitive model for instance segmentation of RSIs.
\end{abstract}


\section{Introduction}

Instance segmentation task means predicting pixel-level results and categories for each instance, and thanks to more accurate results, instance segmentation of remote sensing images(RSIs) has better potential for land planning, urban governance, etc\cite{chen2023mapping}. Instance segmentation based on deep feature learning frameworks makes a great breakthrough and is the mainstream architecture nowadays. For natural scenes such as COCO\cite{lin2014microsoft}, generic instance segmentation models\cite{he2017mask,liu2018path,chen2020blendmask} perform excellently, but when migrating to remote sensing scenes\cite{waqas2019isaid,cheng2016learning}, they endure problems such as complex backgrounds, scale variations, limited instance size, and severe imbalance of foreground and background pixel ratios\cite{chen2021db}.
Although some recent research works\cite{chen2021db,zhang2021semantic,ye2023remote} improve the robustness of the model in terms of scale variation, complex background, etc., the lack of essence understanding and addressing the problems of instance scale limitation and serious imbalance of foreground and background pixel ratio is one of the important reasons for the performance limitation.
\begin{figure}[t]
\centering
\subfloat[]{
\label{fig:dataset}
\includegraphics[scale=0.52]{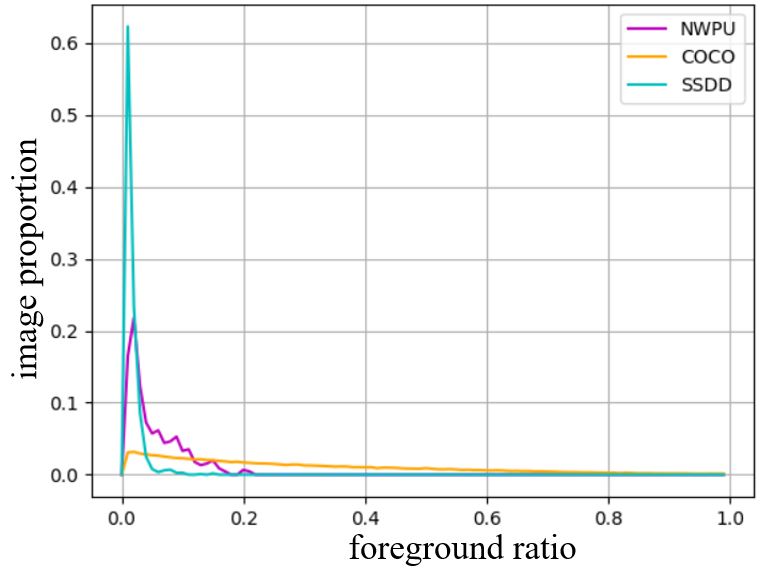}} \\
\subfloat[]{
\label{fig:dataset_samp}
\includegraphics[scale=0.38]{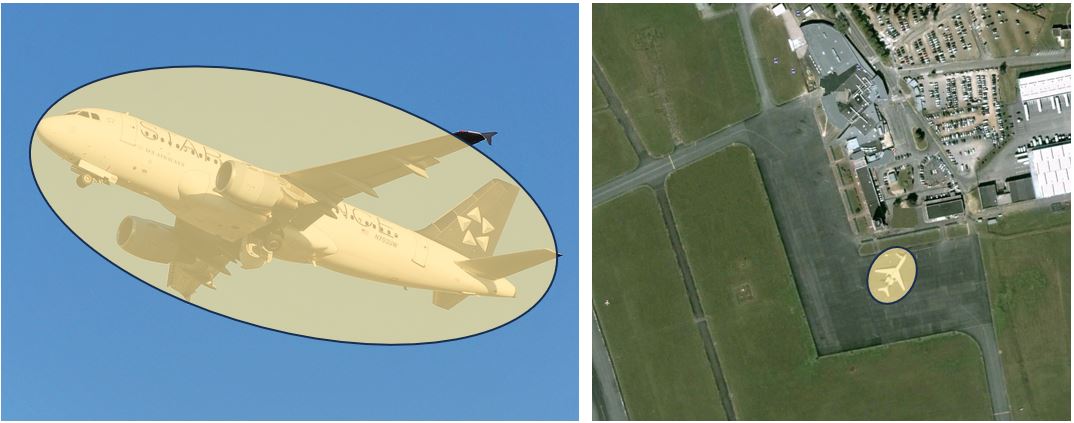}}
\caption{(a) Comparison of image proportion with foreground pixel ratio for natural and remote sensing scene datasets. The foreground pixel ratio of remote sensing scene is much lower than that of natural scene.  (b) The instances in natural scene images and RSIs. The size and foreground pixel ratio of instances in RSIs are pretty lower compared to natural scene images.}
\label{fig_challenge}
\end{figure}

The reasons for the limited performance of remote sensing instance segmentation concerning instance scale constraints and imbalance between foreground and background pixel ratios can be attributed to two aspects: on the one hand, this problem is particularly prominent in remote sensing data in comparison to natural scenes, and on the other hand, some of the operations unique to the current general deep feature learning architectures are detrimental to address this problem. 
For the former, as shown in \autoref{fig_challenge}, the statistical results prove that the imbalance in the ratio of foreground and background pixels and the limited size of the instances in remote sensing images are particularly problematic. The proportion of foreground and background pixels in natural scenes is evenly distributed, whereas for remote sensing scenes, most remote sensing images have no more than ten per cent of foreground pixels, so this problem is more prominent in RSIs.

For the latter, the general feature extraction downsampling paradigm adopted by the existing instance segmentation models is harmful for instance segmentation of RSIs.
As for the existing instance segmentation models\cite{he2017mask,liu2018path,chen2020blendmask} almost all of them are designed based on the backbone network of deep feature learning\cite{he2016deep,liu2021swin}, these contain a large number of downsampling and other operations. 
While this feature extraction paradigm reduces spatial redundancy in natural scenes\cite{lin2014microsoft}, it is detrimental for instance segmentation task of RSIs that possesses few foreground pixel proportions and limited instance size problem.
Because it causes the rich information in the input image not to be fully exploited and utilised.
Therefore, there is an urgent need for a method that can both effectively address this problem in remote sensing data and circumvent the harmful effects of generic deep feature learning downsampling, and then improve the instance segmentation performance. For this purpose, this paper proposes a novel prompt paradigm for remote sensing instance segmentation that takes into account the above two issues together without introducing excessive computation.

\begin{figure}[t]
\centering
\includegraphics[width=3.4in]{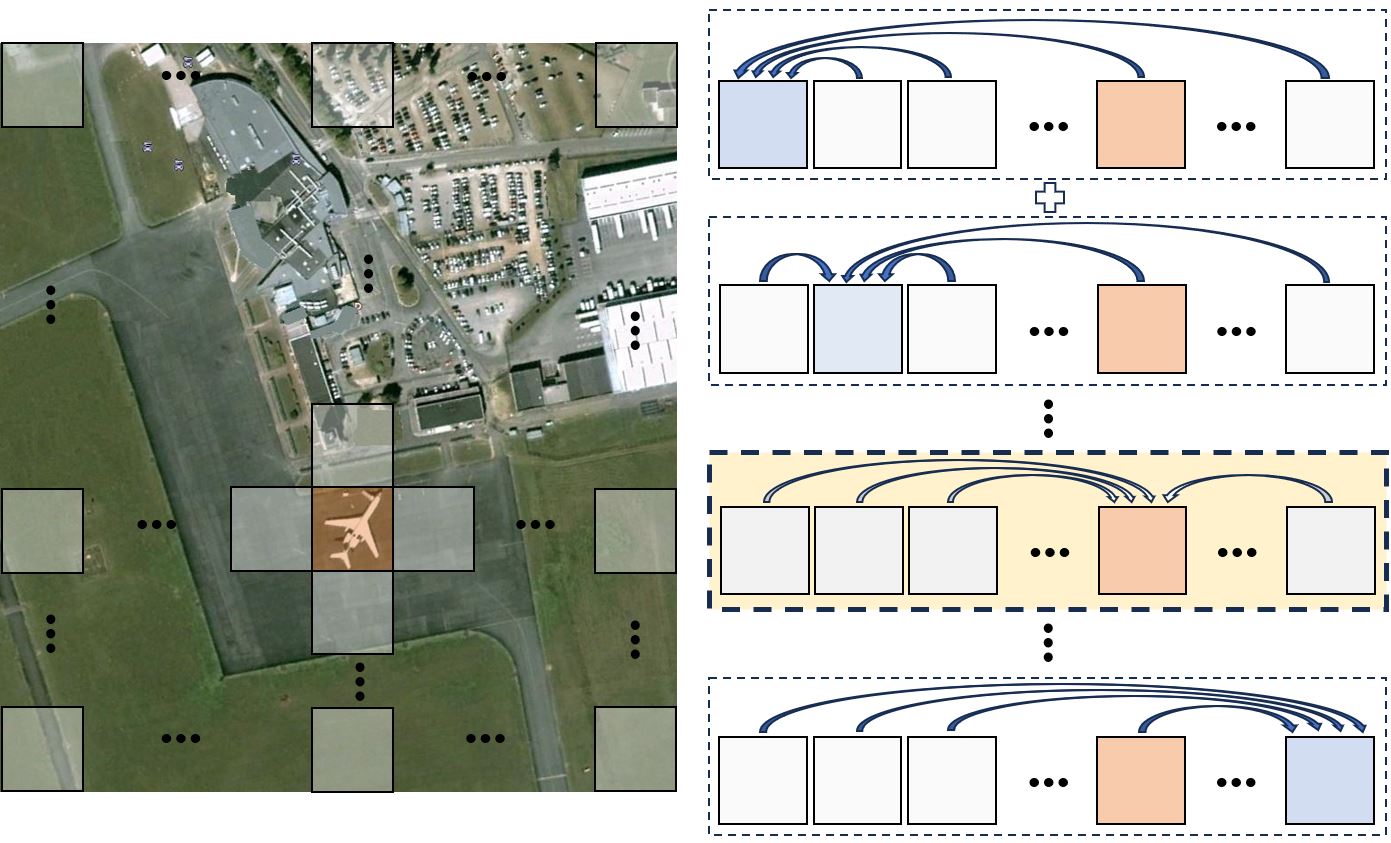}
\caption{The global-to-local idea in our proposed prompt paradigm focuses on modelling the context of global tokens to the local tokens where the instances are located(yellow dotted box), eliminating the enormous computation associated with background tokens interaction(white dotted box).}
\label{fig:idea}
\end{figure}

Prompt learning and transformer have recently demonstrated unique advantages in visual tasks; prompt learning can effectively boost visual downstream tasks with an appropriate prompt template, such as SAM\cite{kirillov2023segment}, and transformer structures are widely adopted due to their powerful representational capabilities, such as VIT\cite{dosovitskiy2020image}. The prompt paradigm proposed in the paper is inspired by the advantages of both and addresses the task characteristics of instance segmentation and the aforementioned issues of remote sensing instance segmentation.
In detail, our prompt paradigm aims to insight the instances in RSIs by appropriate prompt templates, as shown in \autoref{fig:idea}. The paradigm uses a combination of global and local prompt templates designed with transformer structure to mine prompt information from the texture information-rich original image, which in turn addresses the problem of foreground and background imbalance and limited instance size and improves the performance of remote sensing instance segmentation. In particular, the mining is more concerned with the enrichment of the representation of the foreground region where the instances exist that the instance segmentation task is more concerned with, thus allowing for both no downsampling and no redundant computation. Thus, the novel prompt paradigm can also effectively compensate for the harmful effects of generic feature extractor downsampling. It is worth mentioning that the novel prompt paradigm we propose supports the promptable instance segmentation task. 

\begin{figure}[t]
\centering
\includegraphics[width=3.4in]{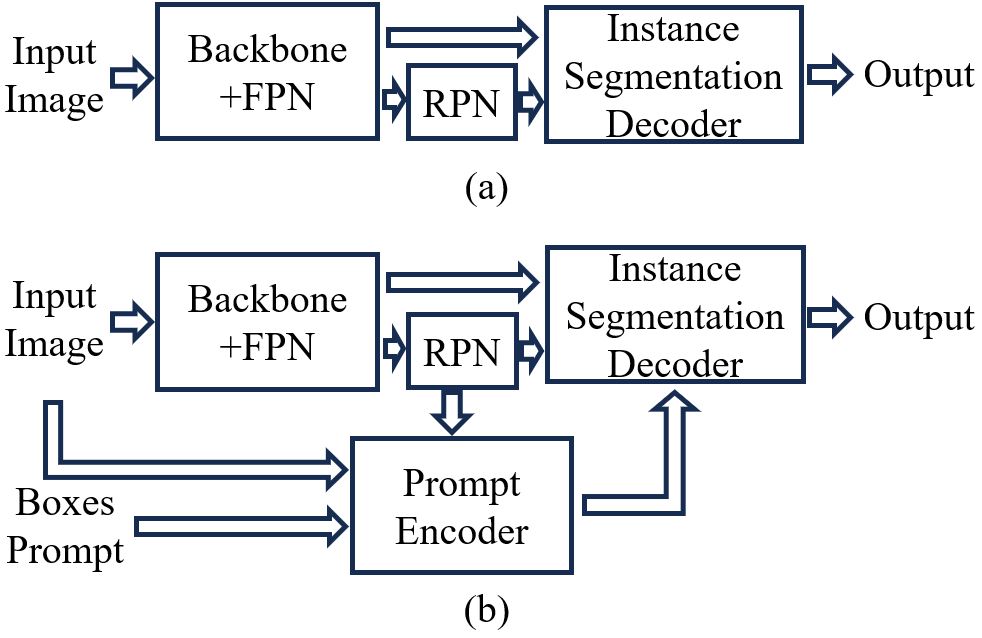}
\caption{Comparison of the instance segmentation with our proposed prompt paradigm and baseline instance segmentation model.}
\label{fig:model_framework_com}
\end{figure}

In essence, our approach consists of three main techniques:
First, a local prompt module is proposed for mining local prompt information in the form of the prompt to boost the representation of specific instances. For a specific instance, the original image contains rich information such as texture and structure, and we design the module without downsampling based on the transformer from a frequency-domain perspective so that the information of different parts of the instance can be fully exploited and interacted. 
Second, a global-to-local prompt module is proposed for mining global prompt information as prompt information for specific instances. This module extracts global-to-local interaction attention features from global tokens for the local tokens in which the instances are located as prompt for a particular instance, as shown in \autoref{fig:idea}.  
Explicitly defining the regions where specific instances are located as local, and then applying the global-to-local idea from the global image to extract the corresponding prompt information, which explicitly enhances the global contextual representation of the specific instances and avoids redundant computation.
The combination of the two prompt modules constitutes the main component of our proposed prompt paradigm, focusing on the fact that the un-downsampled prompt makes it possible to effectively compensate for the problem of limited instance size and imbalance of foreground pixel ratio in RSIs, whereas the instance-specific prompt paradigm and the strong representation module based on the transformer structure make it possible to enhance the representation of the instance objects with instance segmentation focus without excessive computational effort.
Finally, we propose a proposal's area loss function to to optimise the quality of the proposals by adding the decoupling dimension of the scale to the proposals extraction, which in turn better exploits the potential of the above prompt paradigm.  
More valuable is our proposed new prompt paradigm that enables the model to support the promptable instance segmentation task, which we implement by combining the existing instance segmentation models and visual boxes prompt, as shown in \autoref{fig:model_framework_com}.

The contributions in this paper can be summarised as follows:
\begin{enumerate}
\itemsep=0pt
\item{We rethink the problems of front-background imbalance and limited instance size in remote sensing images (RSIs) from both model and data perspective and provide a novel perspective that introduces a novel prompt paradigm to address the above issues.
}
\item{We propose a local prompt module to fully mine the texture information of instances from the input image as a local prompt cue.}
\item{We propose a global-to-local prompt module to model the contextual information from the global tokens to the local tokens where the instances are located as the global prompt cue.}
\item{We conduct thorough evaluation experiments on multiple datasets to prove that our proposed approach is effective for remote sensing image instance segmentation.}
\end{enumerate}

\section{Related work}

\subsection{Instance Segmentation in Remote Sensing Images}

Several works\cite{mou2018vehicle,ye2023remote} in recent years have focused on instance segmentation of remote sensing images.
Mask-RCNN\cite{he2017mask} and Cascade Mask RCNN\cite{cai2018cascade} are two general instance segmentation models, but direct migration to instance segmentation of RSIs limits model performance due to the presence of remote sensing characteristics.
Mou et al.\cite{mou2018vehicle} design a multi-task residual fully-connected convolutional network for vehicle instance segmentation in aerial images.
Wang et al.\cite{wang2023image} propose a cross-domain instance segmentation model based on center feature alignment for domain adaptation of remote sensing building instance segmentation. 
Unlike the above works\cite{mou2018vehicle,wang2023image} that focus only on a single category (e.g., cars, buildings, etc.), the task of segmenting instances considering multiple fine-grained categories is much more difficult.
Chen et al.\cite{chen2021db} propose a balanced strategy for the problem of scale variation and long-tailed distribution in remote sensing instance segmentation.
SCBN\cite{zhang2021semantic} adds a scale complementary mask branch and a semantic attention module constituting a multi-category instance segmentation model to address the problem of complex background and instance scale variations in remote sensing instance segmentation.
Similarly, RSIISN\cite{ye2023remote} incorporates CNN and transformer to design three attention modules of channel-space, multi-scale, and semantic relations to embed the instance segmentation model.

Although these works are effective, their models are all based on deep feature extraction paradigm\cite{he2016deep} and do not realise the harm of operations such as downsampling that is included for the problem of limited instance size or foreground imbalance.
Although HRNet\cite{sun2019deep} recognises the harm and eliminates downsampling, the computation is too large. Therefore, if the downsampling in the deep feature extraction paradigm is removed, the computational effort of the model will be immense, which is not conducive to practical applications.


In view of the above drawbacks, this paper proposes a novel perspective to address the problems of front-background imbalance and low resolution in RSIs and at the same time, circumvent the harms brought by deep feature learning downsampling\cite{he2016deep} to improve the performance of remote sensing image instance segmentation. Briefly, we design a new prompt paradigm where instance-specific prompt information directly mined from input images that have not been spatially downsampled is used as a prompt for decoding the instance segmentation task.
 
\subsection{Prompt Learning in Computer Vision.}
Prompt learning is recently introduced to the computer vision field\cite{jia2022visual,kirillov2023segment} after its remarkable performance in natural language processing field\cite{liu2023pre}.
One application of prompt learning is the vision-language model and fine-tuning of visual model. CoOp\cite{zhou2022learning} uses text as a prompt for visual tasks and aligns text features with image features to improve performance on visual tasks.
VPT\cite{jia2022visual} is a purely visual research work that introduces prompt learning as an embedding form by using learnable vectors as prompt tokens. 
DVPT\cite{ruan2023dynamic} takes into account the instance-specific prompt for vision tasks based on VPT\cite{jia2022visual}, which designs a dynamic instance-wise token for input images.
 
The promptable segmentation task is another research on the application of prompt learning in the visual field. SAM\cite{kirillov2023segment} introduces prompt learning to the segmentation task and constructs a promptable segmentation task that incorporates the construction of large segmentation dataset using prompt learning, as well as the segmentation of manually prompt inputs, achieving significant performance. RSPrompter\cite{chen2023rsprompter} constructs a SAM-based instance segmentation in RSIs by adding an object detection network to extract proposals instead of manually prompt inputs based on the SAM-trained model. Although this achieves some improvement, the number of model parameters is huge, which is not conducive to the deployment of the actual application equipment.


Taken together, the above research works suggest that prompt learning can be applied to various research areas in computer vision, but the core is the design of prompt templates. To address the harms of existing deep feature learning downsampling for RSIs\cite{sun2019deep}, we design a novel prompt template, for instance, segmentation of RSIs. 
Achieving compensation for the harm of downsampling in existing deep feature learning\cite{he2016deep} without significantly increasing computational effort.
Unlike SAM's form of prompts, we mine the prompt information from boxes prompt and original images to use as prompt input.

\begin{figure*}[t]
\centering
\includegraphics[width=7in]{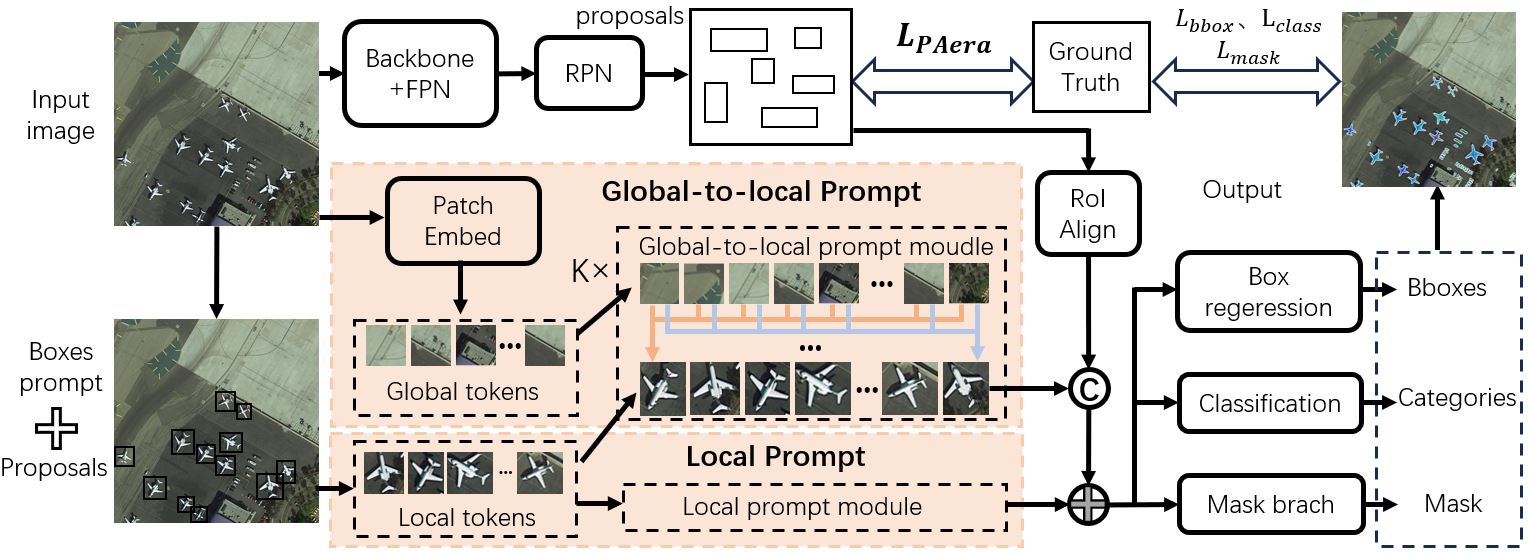}
\caption{The whole framework of our proposed prompt paradigm.}
\label{fig:all_framework}
\end{figure*}
\subsection{Global and Local Attention Mechanism}

Attentional mechanisms are the key to deep learning model network design in recent years, and the combination of global and local attention is one of the popular research works \cite{liu2023survey}. 


VIT\cite{dosovitskiy2020image} is an earlier research work on visual transformer, which divides the input image into tokens by patch embedding, and then computes the correlation information between the image tokens using a deep transformer network containing self-attention to achieve long distance attention.
Focal transformer\cite{yang2021focal} improves the network structure of the transformer to achieve the combination of local and global attention, where one token first performs the computation of short-range local attention with surrounding tokens, and then performs the computation of long-range global attention with tokens that are farther away. SG-Former\cite{ren2023sg} retains the long range dependence of the transformer while redistributing tokens to explore important regions of the image by self-guiding, focusing on information in salient regions while maintaining the global receptive field.
While all of these improvements are valid, they are achieved by stringing together many layers of the network, local and global interactions are implicit, and multiple downsampling is still included in some of these models.
Glass\cite{ronen2022glass} extracts deep features for both the original image and the local image containing text, and then fuses the global and local features to explicitly implement global and local attention. Although such an approach is explicit, the process of extracting depth features multiple times brings a huge amount of computation.

For instance segmentation tasks, the global-to-local attention mechanism can be understood as extracting information from the global image that is useful for local instance object decoding, and thus this global-local attention can be generalised to be instance-specific. From this point of view, this paper designs instance-specific global-local attention.
Our proposed approach enables explicit global-local attention based on the transformer structure, while using a form of prompt learning without introducing enormous parameters and computations, as shown in \autoref{fig:idea}.





\section{Methodlogy}
This section details the methodological details of our proposed approach. The issues of remote sensing image instance segmentation is rethought and modelled by the paper in \autoref{sec:modeling}. The \autoref{sec:GloabelToLocal} details several major innovations in our proposed prompt paradigm. In \autoref{sec:LPM}, the local prompt module is detailed. In \autoref{sec:GPM}, the global-to-local prompt module used for specific instances is detailed. The \autoref{sec:PAreaLoss} describes the design of the proposals's area loss function. In \autoref{sec:promptableIS}, The implementation of the promptable instance segmentation task constructed based on our proposed prompt paradigm is detailed.  

\subsection{Rethink the Instance Segmentation in Remote Sensing Images}
\label{sec:modeling}
Although the deep feature extraction paradigm extracts rich semantic information of deep features, the operation of multiple downsampling makes the spatial detail information lost, which exacerbates the problem of front-background imbalance and limited instance size of remote sensing images.
Given the input image $I$, the $i$ th stage $F_{i}$ of the generic deep feature extraction paradigm can be formulated as:
\begin{equation}
\begin{aligned}
F_{i} &= e^{\beta\cdot i}\cdot e^{-\alpha\cdot s_{i}^{2}}\cdot I
\end{aligned}
\end{equation}
where the information gain in the channel dimension is modelled as $e^{\beta\cdot i}$, $\beta$ is the gain coefficient. And the information decay in the spatial dimension can be modelled as $e^{-\alpha\cdot s_{i}^{2}}$. $\alpha$ is the decay coefficient. $S_{i}$ denotes the number of down-sampled times of the ith stage in the spatial dimension.

And the existing instance segmentation model based on above deep feature extraction paradigm can be formulated as: 
\begin{equation}
\begin{aligned}
R &= D(\sum_{i=0}^{M}F_{i})=D(\sum_{i=0}^{M}e^{\beta\cdot i}\cdot I\cdot e^{-\alpha\cdot s_{i}^{2}})
\end{aligned}
\end{equation}
where $D(\cdot)$ indicates the instance segmentation decoder. The decoding of instance segmentation relies on the output of deep feature extraction $\sum_{i=0}^{M}F_{i}$, $M$ being the number of multi-scale, and because of the information decay in the spatial dimensions thereof, the result of instance segmentation $R$, is to some extent limited by this decay. 
Although some research works\cite{sun2019deep} try to eliminate this information decay $e^{-\alpha\cdot s_{i}^{2}}$ by removing the operation of downsampling in the deep feature extraction paradigm, which can be modelled as $\sum_{i=0}^{M}F_{i} = \sum_{i=0}^{M}e^{\beta\cdot i}\cdot I$, the excessive number of layers of the backbone network in the deep feature extraction paradigm brings about excessive computational and parameters, which is not conducive to practical applications. 

In light of these drawbacks, we address these issues with prompted learning. Overall, we design the novel prompt paradigm and construct new prompt encoder to encode the input prompt, which in turn participates in instance segmentation decoding. Given the input image, the prompt encoder in our proposed prompt paradigm of instance segmentation in RSIs can be described as:
\begin{equation}
\begin{aligned}
P &= e^{\gamma}\cdot I
\end{aligned}
\end{equation}
where $e^{\gamma}$ models the dimensional information gain. $\gamma$ is the gain coefficient.
Our proposed prompt encoder is neither spatial information decay nor over parameters and computation, because our proposed prompt paradigm considers the task-specific characteristics of instance segmentation, the use of a small number of strong representational structures, and so on. The instance segmentation model based on our proposed prompt paradigm can be described as:
\begin{equation}
\begin{aligned}
R &= D(\sum_{i=0}^{M}F_{i}, P)\\
&=D(\sum_{i=0}^{M}e^{\beta\cdot i}\cdot I\cdot e^{-\alpha\cdot s_{i}^{2}}, e^{\gamma}\cdot I)
\end{aligned}
\end{equation}

Our proposed prompt paradigm encodes the input prompt to participate in the instance segmentation decoding process in a cueing manner, as shown in \autoref{fig:all_framework}, which retains the advantages of deep feature extraction and compensates for the decay of spatial information due to downsampling, as the prompt encoding avoids the decay of spatial information and preserves the general instance segmentation model.

\subsection{Global to Local Prompt for Instance Segmentation}
\label{sec:GloabelToLocal}
Existing instance segmentation models are almost always designed based on deep feature extraction, which contains downsampling that is detrimental to addressing the problem of limited instance size and unbalanced foreground pixels in RSIs. Therefore, we propose an approach to compensate for this deficiency and address the problem of remote sensing images in a visual prompt paradigm. Our proposed prompt paradigm adopts the idea of combining local and global to design, which mainly contains two parts: insight local prompt information and insight global prompt information.

\subsubsection{Local Prompt Module for Specific Instances}
\label{sec:LPM}
For instances in remote sensing images, there are problems of limited size and low foreground pixel ratio, which makes the feature extraction difficulty for instances more difficult, coupled with the detriment of downsampling in depth feature extraction, there is an urgent need to compensate for these to insight the representation of the instances, and thus improve the performance of instance segmentation in RSIs. Therefore, we propose a local prompt module for specific instances (LPM), aiming to exploit the rich texture information of the local regions where the instances are located in the original image in a prompt manner. 
\begin{figure}[t]
\centering
\includegraphics[width=3.4in]{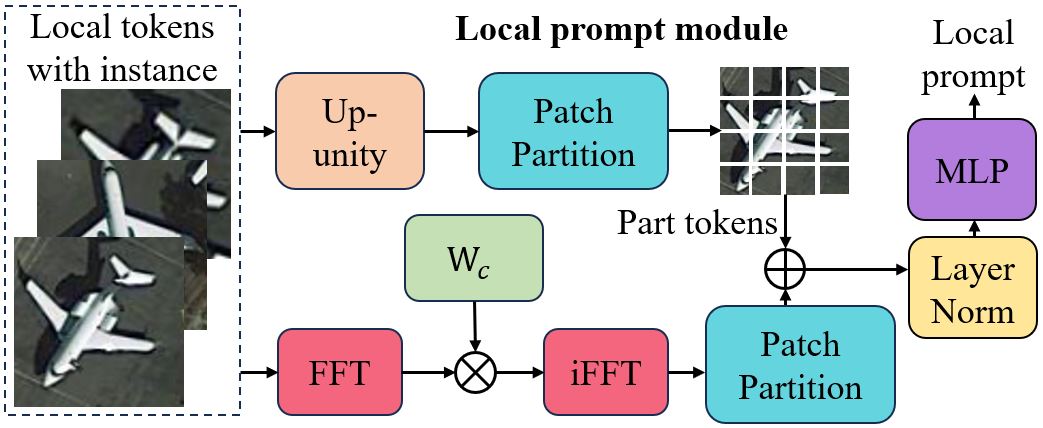}
\caption{The design detail of our proposed local prompt module. Up-unity means up-sampled and adjusted to a unified size.}
\label{fig:local_prompt}
\end{figure}

The local prompt module is designed as follows. First, the definition of local in the local prompt module is the local regions where instances exist in the RSIs. These local regions in the original images are cropped as inputs to the LPM, and are divided into tokens with the overlapping patch partition. And the different parts of the instance are contained in these tokens. The spatial location coordinates of these local regions are obtained from the inputs of the box prompts and the proposals. Specially, we add an interaction from a frequency domain perspective to enhance the representation of the tokens where different parts are located. Parallel to this, we the fast fourier transform to transform these cropped local regions, add learnable parameter embeddings, and then inverse fast fourier transform, after patch partition divided into tokens and concated with part tokens, and then MLP layer computation of interactions to get the local prompt.

Another benefit of our proposed form of LPM to implement the above modelling idea is that it can improve the robustness of the model while achieving the above modelling goals. The reason is that the prompt information extracted by the LPM is extracted at the instance level, and the inputs only correspond to the regions where the instance objects are located, without including too much background region, which can effectively avoid the interference of the background. This contributes to the robustness of the model.

\subsubsection{Global-to-local Prompt Module for Specific Instances}
\label{sec:GPM}
For small-sized instances in remote sensing images, their own texture information is lacking, so improving their representation through global contextual information is necessary. Unlike existing feature extraction networks that have a transformer structure with a large number of layers such as focal transformer\cite{yang2021focal}, we build a relatively lightweight global-to-local transformer prompt network to insight instance-specific context modelling, named global-to-local prompt module for specific instances (GPM). 
While existing transformer networks are aware of the idea of global-to-local modelling, the ambiguity of the definition of local leads to the need for a large number of network layers in order to achieve this modelling, whereas our proposed GPM explicitly defines the local regions where instances are located in the RSIs as local that needs to be insighted with global information, as shown in \autoref{fig:all_framework}.
\begin{figure}[t]
\centering
\includegraphics[width=3.4in]{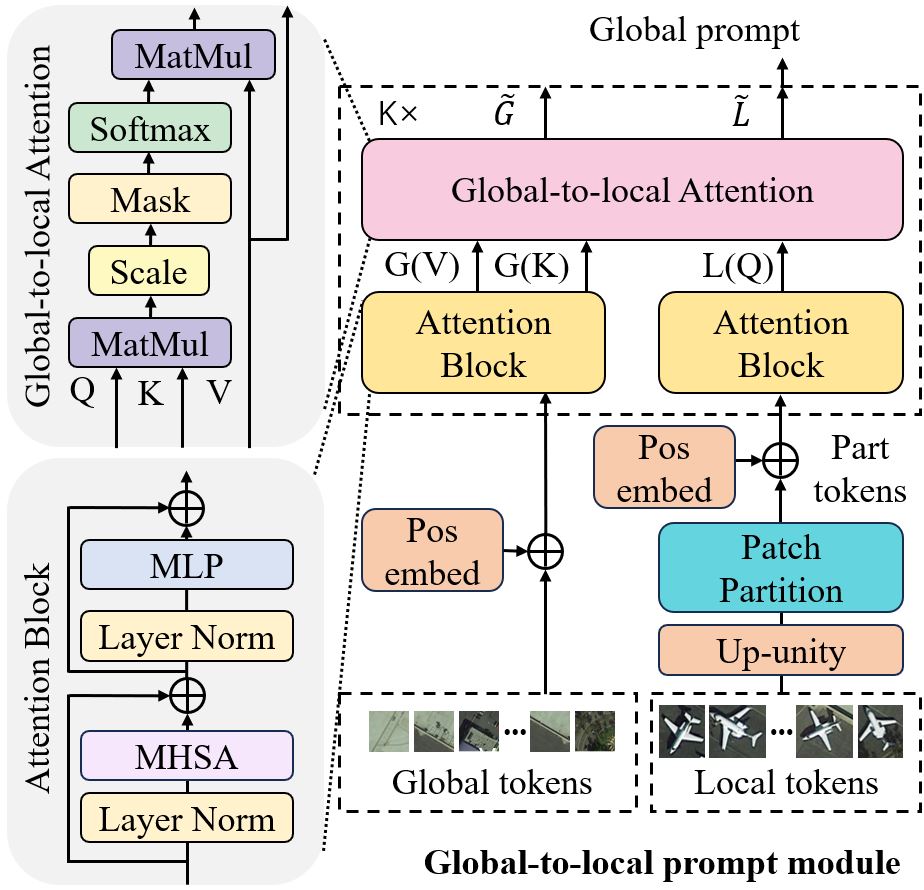}
\caption{The design detail of our proposed global-to-local prompt module.}
\label{fig:g2l_prompt}
\end{figure}

The detailed design of the GPM is shown in \autoref{fig:g2l_prompt}. For the original image, we use the patch partition to divide it into global tokens as partial input to GPM. For the other partial of the input: the local tokens where the instances are located, GPM up-samples them to the same size, and then divides them into part tokens using a patch partition. GPM adds positional embeddings to both global tokens and part tokens, and then the multi-head self-attention module in the transformer structure performs its own interaction, denoted as $G$ and $L$. Then global-to-local attention computation $G2LAttn(\cdot)$ is performed to obtain local tokens that aggregate global context information $\tilde{L}$, as indicated in \autoref{equ:g2l}. Such self-interactions and global-to-local computational loops are performed multiple times.

\begin{equation}
\begin{aligned}
\tilde{L} = G2LAtt(G,L) = softmax(\frac{LG^{T}}{\sqrt{d_k}})\ast G\\
\end{aligned}
\label{equ:g2l}
\end{equation}

In terms of model size and practical application, the GPM introduces a small number of parameters in the form of prompt learning to achieve global contextual information extraction for instance-specific objects, which has obvious advantages over the large-scale parameter learning in existing visual transformer works\cite{yang2021focal,ren2023sg}. The core advantage is that for the characteristic of instance segmentation task where the instance region representation is the core, GPM focuses on modelling the global contextual representation of the instances, thus eliminating redundant interactions between background regions. as shown in \autoref{fig:idea}.

The local prompt module and the global-to-local prompt module form the core of our proposed prompt paradigm. The cleverness of this paradigm is that it is instance-specific, with the advantage of both insighting the representation of the instance and avoiding redundant computational effort in existing transformer feature extraction networks, as shown in \autoref{fig:idea}.

\subsubsection{Proposals's Area Loss Function}
\label{sec:PAreaLoss}
While the implementation of the LPM and the GPM described above can effectively utilise the potential of prompt learning in remote sensing image instance segmentation, the implementation of these two modules partially relies on the proposals, which means that the quality of the proposals determines the performance of the above modules with a certain probability. In order to obtain more accurate proposals, we propose a loss function that makes the proposals more accurate in 2D spatial scale, named proposals's area loss function (PAreaLoss). The details of the design are described below.

The process of proposal extraction in existing models can be summarised as the decoupling of positioning and correction. Our proposed PAreaLoss adds a scale of decoupling dimensions, which allows the decoupling of the location, scale, and correction of the instance segmentation model for instance boundary regression.
In the instance segmentation model of joint object detection, the region proposal network initially locates some anchors that may contain instances as proposals from a large number of preset anchors, after which the detection branch and mask branch perform boundary correction to get the final result\cite{he2017mask}. However, the scale difference between instance categories of RSIs is large, so the scale factor needs to be taken into account.

During model training, the PAreaLoss calculates the bias of the model's predicted proposals and ground boxes in terms of scale area. The iou value is used as a criterion to assign the proposals to the ground-boxes, and after the matching is done, the bias of the area between the corresponding proposal and the ground-box is calculated and normalised, and finally averaged. Proposals characterise spatial location information of foreground instance objects, so the addition of the scale decoupling constraints allows proposals to represent the spatial location of instances more accurately, avoiding too much background noise or missing parts.
The detailed calculation process of the is described in Algorithm \ref{algorithm1}. 
\begin{algorithm}[h]
\caption{Calculation of the proposals's area loss function}
\label{algorithm1}
\begin{algorithmic}[1]
\Require{input image $I$; predicted proposals $Proposals_{N}$; GT-boxes $gtBoxes$.}
\Ensure{proposals's area loss $L_{PArea}$.}\\
\emph{Initialization:}\\
\emph{$W,H \gets I.size.width, I.size.height$}\\
\emph{$W_{gt},H_{gt} \gets gtBoxes.size.width, gtBoxes.size.height$}\\
\emph{Uniform scale between predicted proposals and GT-boxes:}\\
\emph{$Scale_W, Scael_H \gets W_{gt}/W, H_{gt}/H$}\\
\emph{$Proposals_N[:,x_{min}] \gets Proposals_N[:,x_{min}]*Scale_W$}\\
\emph{$Proposals_N[:,y_{min}] \gets Proposals_N[:,y_{min}]*Scale_H$}\\
\emph{$Proposals_N[:,w] \gets Proposals_N[:,w]*Scael_W$}\\
\emph{$Proposals_N[:,h] \gets Proposals_N[:,h]*Scael_H$}\\
\emph{Calculate the loss function:}
\For {$i, proposal$ in enumerate($Proposals_{N}$)}
\State$gtBox \gets argmaxIoU\big(gtBoxes,proposal\big)$
\State$area_{proposal} \gets porposal.w * proposal.h$
\State$area_{gt-box} \gets gtBox.w * gtBox.h$
\State$area_{error} \gets ABS(area_{proposal}-area_{gtbox})$
\State$L_{i} \gets \big(area_{error}/(area_{gtbox} + eps)\big)^{2}$
\State$Proposals_{N}.pop(proposal)$
\EndFor\\
\emph{Averaging loss values of multiple maps:}\\
\emph{$L_{PArea} \gets \frac{1}{N}\sum_{i=1}^{N} L_{i}$}\\
\Return $L_{PArea}  $
\end{algorithmic}
\end{algorithm}

All in all, the advantage of adding the decoupling of the proposals scale factor is that the model's processing of proposals becomes three parts: localisation, scaling, and correction, which makes the extracted proposals more precise and more conducive to better performance of our proposed LPM and GPM. The whole loss function during training phase $L_{total}$ can be described as:
\begin{equation}
\begin{aligned}
L_{total} = L_{PArea} + L_{box} + L_{class} + L_{mask}
\end{aligned}
\end{equation}
where $L_{box}$, $L_{class}$, and $L_{mask}$ denote the loss functions for the box regression, classification, and mask branches of the instance segmentation decoder, respectively.
\subsection{Promptable Instance Segmentation}
\label{sec:promptableIS}
Much recent works focus on visual promptable tasks, such as the promptable segmentation task proposed by SAM\cite{kirillov2023segment}. These research efforts show that prompting visual downstream tasks can contribute to performance on visual downstream tasks such as segmentation. Therefore, we construct a new model using the above proposed components that can support the promptable instance segmentation task to achieve interactive instance segmentation. The implementation consists of two parts, the design of the prompt process and the optimisation of the training.

We design the process for promptable instance segmentation with the proposed components based on the existing instance segmentation model. Similar to the visual prompt proposed by SAM\cite{kirillov2023segment}, we adopt box prompt as the main prompt input for instance segmentation of RSIs, but the difference is that the boxes prompt is derived from two parts, one part is the proposals output from the region proposal network\cite{he2017mask}, and the other part is the manually box prompt. The form of manual box prompt allows our proposed promptable instance segmentation model to support interactive instance segmentation task, while proposals output by the region proposal network as box prompt allow end-to-end automated instance segmentation models to also take advantage of our proposed prompt paradigm to improve performance of instance segmentation. With the input of boxes prompt, our above proposed LPM and GPM act as prompt encoder to encode the box prompt as deep semantic features that participate in the decoding process of the instance segmentation decoder, which in turn yields the instance segmentation mask results.

We optimise the training process to better achieve promptable instance segmentation. During training, we construct the training process as an optimisation process similar to the interactive promptable instance segmentation. Specifically, in the forward propagation of each iter in training, the proposals output from region proposal network and randomly selected part of the ground-boxes are used as inputs to the instance segmentation decoder, and then roi pooling operation is carried out to get the corresponding semantic feature blocks; they are also used as inputs to the local positional coordinates of the LPM and the GPM, and after prompt encoding to get the prompt information, and the semantic feature blocks and the prompt information are jointly involved in the mask prediction of instance segmentation.
In addition, a small range of coordinate random noise is added to the ground-boxes used as the box prompt, making a better fit with the manual box prompts.

The above two improvements allow our proposed prompted instance segmentation model to both enable interactive prompted instance segmentation and allow existing end-to-end automatic instance segmentation models to better exploit our proposed prompt paradigm to improve the performance of instance segmentation.

\section{Experiments}

In this section, we set up evaluation experiments to evaluate the effectiveness of our proposed approach from multiple perspectives. Intuitively, our proposed optimised instance segmentation model for the remote sensing image characterisation problem is evaluated on multiple datasets, based on several existing models.

\subsection{Datasets}
We adopt four popular remote sensing instance segmentation datasets to construct experiments to illustrate the effectiveness of our proposed approach, which include two optical remote sensing image instance segmentation datasets (ISAID and NWPU VHR-10) and two SAR image remote sensing instance segmentation datasets (SSDD and HRSID).

\begin{table}
\footnotesize
\renewcommand{\arraystretch}{1.5}
\tabcolsep=0.30cm
\centering
\caption{Image proportions for different foreground pixel ratios of instances in remote sensing scene and natural scene. }
\label{tab:com_dataset}
\begin{tabular}{c|c|ccc}
\cline{1-5}
\multicolumn{2}{c|}{foreground ratio}  & [0.0,0.1) & [0.1,0.2) & [0.2,1.0] \\                              
\cline{1-5} 
natural scene & COCO    & 24.5\% & 20.1\% & 55.4\% \\\cline{1-5}
\multirow{2}{*}{\makecell[c]{remote sensing\\scene}} & NWPU          & 84.1\% & 14.8\% & 1.1\% \\ \cline{2-5}
         & SSDD          & 99.5\% & 0.5\% & 0.0\% \\ 
\cline{1-5}
\end{tabular}
\end{table}

ISAID. The ISAID dataset\cite{waqas2019isaid} is a large instance segmentation dataset of remote sensing optical images, which contains 655,451 target instances. The dataset provides bounding box labelling, segmentation boundary labelling, and fine-grained categories for each target instance, with a total of 15 fine-grained categories for airplanes, ships, storage tanks, baseball fields, tennis courts, basketball courts, ground track and field, ports, bridges, large vehicles, small vehicles, helicopters, roundabouts, swimming pools, and football fields. The original remote sensing image is cropped to 800*800 size and contains 28,029 training set, 9,512 validation set and 19,377 test set where the labelling of the test set is not disclosed. 
The ISAID contains a large number of scale-limited instances, with 52 per cent of instances with fewer than 144 pixels and 85.7 per cent of instances with fewer than 1024 pixels. The massive number of target instances is sufficient to support the validation of the effectiveness of our proposed approach, and thus ISAID serves as the primary dataset for constructing the experiments.

NWPU VHR-10. The NWPU VHR-10 dataset\cite{cheng2016learning} we adopt is an extension of the NWPU WHR-10 object detection dataset for instance segmentation, which contains 650 remote sensing optical images. The instance categories of annotations include aircraft, ships, oil tanks, baseball fields, tennis courts, basketball courts, athletic fields, harbours, bridges, and automobiles for a total of 10 categories. As shown in \autoref{tab:com_dataset}, we performed a statistical analysis based on the annotations provided by the dataset, which shows that 98.9\% of the image has less than 20\% of instance foreground pixels, which is significantly lower than the COCO dataset of natural scenes. In our experiments, we split the training and validation sets in the ratio of 7:3.

SSDD. The SSDD dataset\cite{zhang2021sar} is a SAR image interpretation dataset containing 1160 SAR images under a variety of conditions, cropped to a resolution of 500*500. The manual annotation used for the instance segmentation task is adopted in experiments, which contained 2456 ships. The statistics in \autoref{tab:com_dataset} show that 99.5 per cent of the images have a ratio of instance foreground pixels of no more than 10 per cent, and the problem of unbalanced foreground pixels is more prominent. Similarly, we adopt the 7:3 ratio to split the training set and the validation set.

HRSID. The high resolution sar images dataset\cite{wei2020hrsid} is an SAR image interpretation dataset that supports ship detection, segmentation, and instance segmentation tasks, and contains 5604 SAR images with different resolutions, polarisations, and sea areas. The resolution of the images ranges from 0.5m to 3m. The dataset provides an instance segmentation annotation containing 16,951 ship instances. 

\subsection{Evaluation Metrics and Implementation Details}
Evaluation metric.
As with most instance segmentation works\cite{vu2021scnet,ye2023remote}, the paper adpot the COCO evaluation metric to verify the model performance. 
$AP$, $AP_{50}$, $AP_{75}$ are the evaluation metrics to evaluate the performance of the model on the overall dataset. $AP$ (mean average precision) value is calculated by category-averaged mean precision of iou ranging at 0.5-0.95 in 0.05 step.
And $AP_{50}$, $AP_{75}$ means the mean average precision value at Iou threshold of 0.5, 0.75.
$AP_{s}$, $AP_{m}$, $AP_{l}$ are the evaluation metrics for evaluating the performance of the model on small-scale instance set, medium-scale instance set, and large-scale instance set, and we also adopt these three evaluation metrics to illustrate the impact of our proposed approach on the performance of different scale instances.
In addition, the object detection task and the instance segmentation task are interrelated tasks in the instance segmentation model, and we denote the performance of the two tasks by $AP^d$ and $AP^s$, respectively.

Implementation details.
All the experiments in the paper are conducted on the MMdetection framework\cite{chen2019mmdetection}. The weight initialization of the ResNet50 and ResNet101 adopts pre-trained weights in ImageNet dataset\cite{deng2009imagenet}. The models are trained on two 3090 GPUs with the initial learning rate preset to 0.02. The hypermeter $S_{ps}$ is preset to 28 and the determined experiments are detailed in \autoref{sec:ablation_img_roi_size}. The size of the image input is fixed at 800*800 for both training and inference. In all the experiments of training and testing, the experiments of our proposed method did not use any tricks such as multiscale training, multiscale inference, and so on.

\subsection{Ablation Studies}
\begin{figure}[t]
\centering
\includegraphics[width=3.4in]{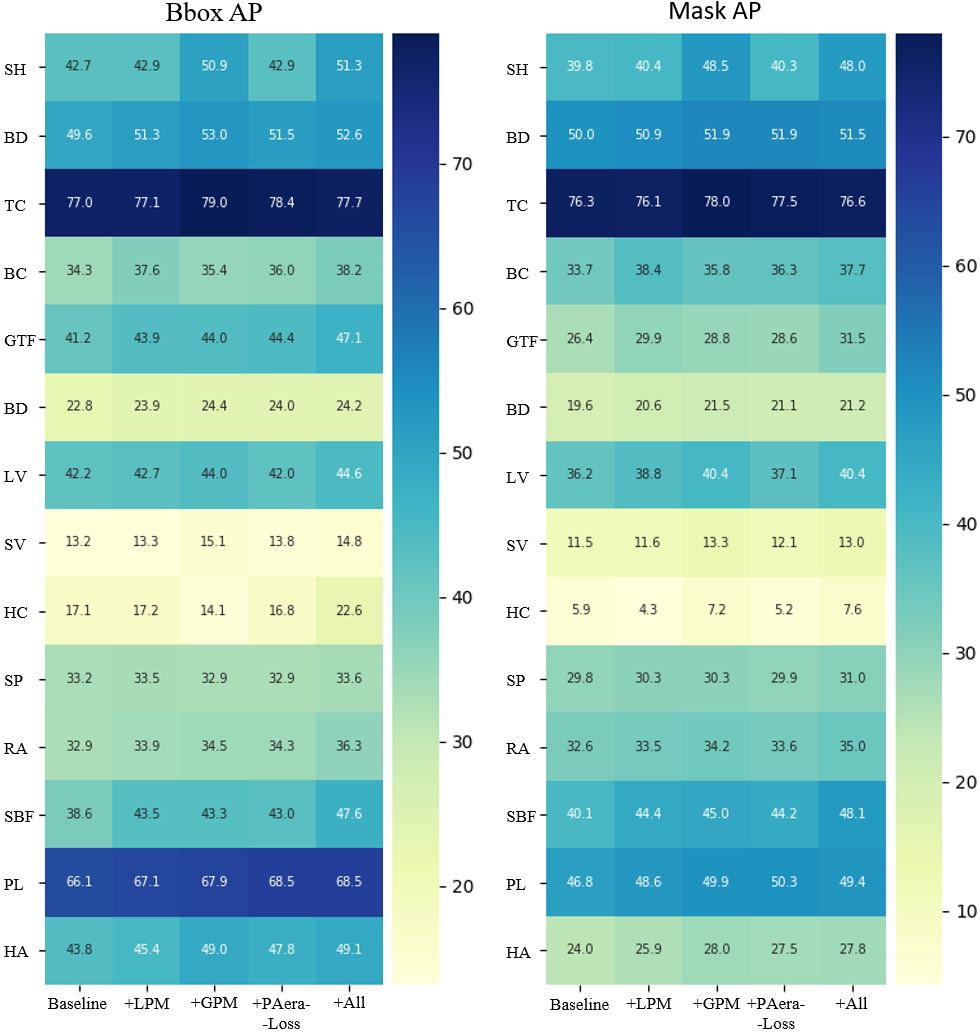}
\caption{Ablation experiments in fine-grained categories.}
\label{fig:ablation_category}
\end{figure}

In this section, we build ablation experiments to illustrate the effectiveness of the different components of our proposed approach from multiple perspectives, including performance on the overall dataset (\autoref{tab:abla}), performance on fine-grained categories (\autoref{fig:ablation_category}), and hyper-parameterised ablation (\autoref{fig:ablation_img_roi_size}).

\begin{table*}
\footnotesize
\renewcommand{\arraystretch}{1.5}
\centering
\caption{Ablation studies of each component in our approach. All ablation experiments are conducted on the validation set of the ISAID dataset. }
\label{tab:abla}
\begin{tabular}{c|c|ccc|ccc|ccc|c|c}
\cline{1-13}
 \multirow{2}{*}{Model} & \multirow{2}{*}{Backbone} & \multirow{2}{*}{LPM} & \multirow{2}{*}{GPM} & \multirow{2}{*}{PAreaLoss} & \multicolumn{3}{c|}{Bbox $AP$} & \multicolumn{3}{c|}{Mask $AP$}  & \multirow{2}{*}{FPS} & \multirow{2}{*}{Model Size} \\\cline{6-11}                             
  & & & & & $AP$ & $AP_{50}$ & $AP_{75}$ & $AP$ & $AP_{50}$ & $AP_{75}$ & & \\
\cline{1-13} 
 \multirow{5}{*}{Mask-RCNN\cite{he2017mask}} &\multirow{5}{*}{ResNet50} & & & & 39.6 & 61.4 & 43.6 & 33.8 & 56.6 & 35.7 & 16.8&    44.1M  \\ 
 & & $\surd$&       &           & 40.9 & 62.9 & 45.2 & 35.3 & 58.0 & 37.3 & 15.5&  46.4M      \\ 
 & &        &$\surd$&           & 41.3 & 62.8 & 46.9 & 36.0 & 59.2 & 40.0 & 14.9&  49.2M  \\ 
 & &        &     &     $\surd$ & 41.2 & 62.1 & 46.0 & 35.4 & 57.5 & 38.0 & 16.8& 44.1M \\ 
 & &$\surd$ &$\surd$   &$\surd$ & 42.8 & 64.6 & 48.3 & 36.5 & 60.4 & 39.9 & 13.1 & 51.3M    \\
\cline{1-13}
\end{tabular}
\end{table*}

\subsubsection{Effect of local prompt module (LPM)}

The local prompt module (LPM) is proposed to better exploit the texture detail information of the instance objects in the local areas from the original image to be used as the prompt information for instance segmentation decoding, which in turn improves the performance of instance segmentation. The second row of \autoref{tab:abla} shows the results of the baseline model with the addition of this module, which achieves a significant improvement in both detection and segmentation performance compared to the baseline model in the first row. For object detection it improves $AP^{d}$: 1.3\%, $AP^{d}_{50}$: 1.5\%, $AP^{d}_{75}$: 1.6\% and for segmentation it improves $AP^{s}$: 1.5\%, $AP^{s}_{50}$: 1.4\%, $AP^{s}_{75}$: 1.6\%. 
The second column of \autoref{fig:ablation_category} lists the performance of the fine-grained categories of the model with the LPM embedded. Compared to the baseline model in the first column, our proposed LPM delivers performance improvements on the vast majority of categories, especially for BC and SBF, with instance segmentation and object detection improvements of $AP^{s}: $4.7\%, $AP^{s}$: 4.3\% and $AP^{d}$: 3.3.\%, $AP^{d}: $4.9\% respectively. 
Achieving such a significant improvement proves that our proposed LPM is indeed able to better exploit the texture detail information of the instances from the original image, and such a prompt are also beneficial for instance segmentation in remote sensing images. In short, our proposed LPM is effective for addressing the problem of remote sensing instance segmentation. 

\subsubsection{Effect of global prompt module (GPM)}

The global prompt module (GPM) is designed to exploit global contextual information to the target instances from the un-downsampled original image, which in turn improves the model's performance for instance segmentation in the form of prompt cues. The third row of \autoref{tab:abla} shows the performance of the model with the GPM embedded. Compared to the baseline model in the first row, our proposed GPM improves $AP^{s}$: 2.2\%, $AP^{s}_{50}$: 2.6\%, $AP^{s}_{75}$: 4.3\% in instance segmentation; and $AP^{d}$: 1.7\%, $AP^{d}_{50}$: 1.4\%, $AP^{d}_{75}$: 3.3\% in object detection.
The third column of \autoref{fig:ablation_category} shows the model performance of fine-grained categories with the GPM embedded. Compared to the baseline model in the first column, our proposed GPM achieves an improvement of $AP^{s}$: 8.7\%, $AP^{d}$:8.2\% and $AP^{s}$: 4.0\%, $AP^{d}$: 5.2\% in the SH and HA categories, respectively, and in most of the other categories to varying degrees. 
From the analysis, our proposed GPM contains global-to-local prompt paradigm ideas and designed implementation modules that bring different levels of enhancement in overall data and fine-grained categories. In short, our proposed GPM is effective for remote sensing image instance segmentation.

\subsubsection{Effect of proposal's area loss function (PAreaLoss)}

The proposals's area loss function (PAreaLoss) is proposed to to add a scale decoupling dimension for the extraction of proposals to achieve more accurate proposals, which in turn exploits the potential of our proposed LPM and GPM to achieve better performance.
Compared to the baseline model in the first row of \autoref{tab:abla}, the fourth row with the addition of PAreaLoss has an improvement in instance segmentation performance: $AP^{s}$: 1.6\%, $AP^{s}_{50}$: 0.9\%, $AP^{s}_{75}$: 2.3\%, and an improvement in object detection performance: $AP^{d}$: 1.6\%, $AP^{d}_{50}$: 0.7\%, $AP^{d}_{75}$: 2.4\%. Compared to the baseline (first row), our proposed three components (fifth row) achieve an improvement in instance segmentation: $AP^{s}$: 2.7\%, $AP^{s}_{50}$: 3.8\%, $AP^{s}_{75}$: 4.2\% and an improvement in object detection: $AP^{d}$: 3.2\%, $AP^{d}_{50}$: 3.2\%, $AP^{d}_{75}$: 4.7\%. In the ablation of the fine-grained categories in \autoref{fig:ablation_category}, PAreaLoss improves to varying degrees in most of the categories, especially for SBF, instance segmentation performance improves by $AP^{s}$: 4.1\%, and object detection performance improves by $AP^{d}$: 4.4\%. From the analyses, our proposed PAreaLoss indeed improves the performance of instance segmentation due to its benefit of more accurate proposals, which in turn exploits the potential of LPM and GPM. In short, our proposed PAreaLoss is effective for remote sensing image instance segmentation.

\subsubsection{Hyperparameter of prompt roi size $S_{ps}$}
\label{sec:ablation_img_roi_size}
\begin{figure}[t]
\centering
\includegraphics[width=3.4in]{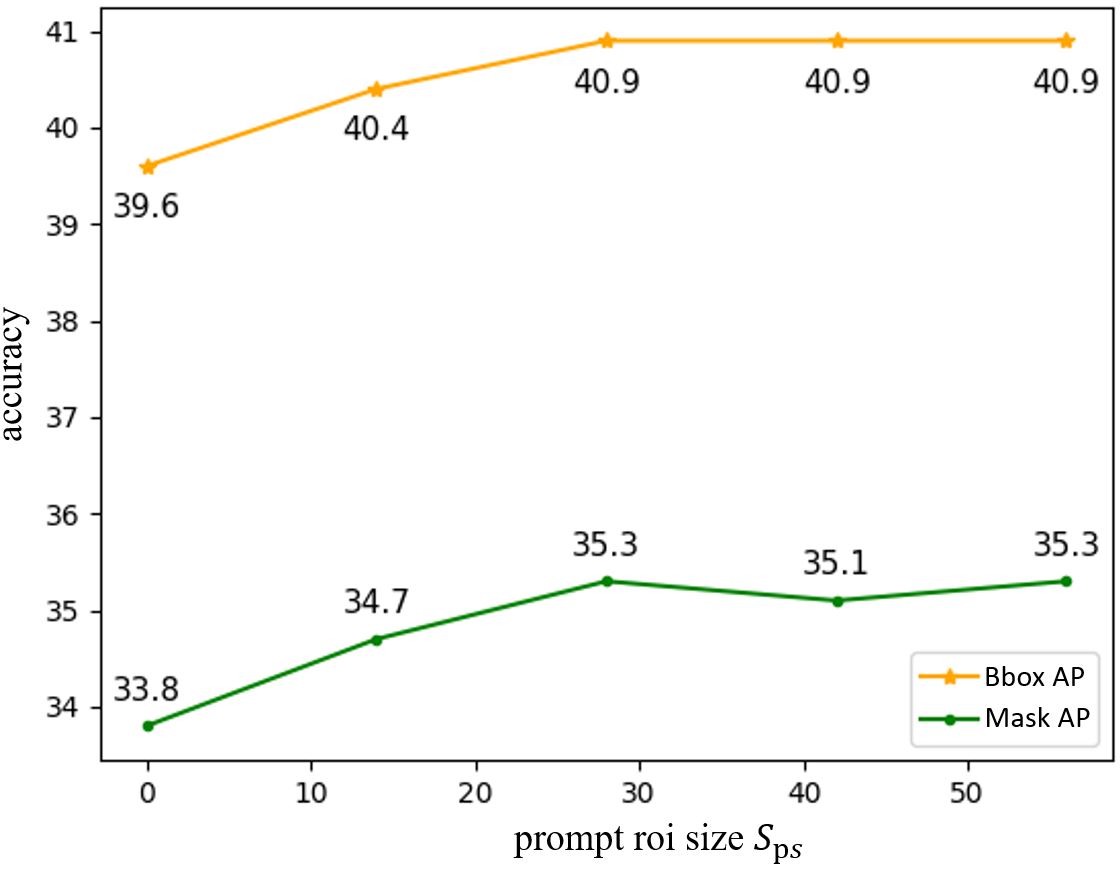}
\caption{The albation of the hyper-parameter prompt roi size $S_{ps}$. }
\label{fig:ablation_img_roi_size}
\end{figure}
As shown in \autoref{fig:ablation_img_roi_size}, the ablation experiments of the hyperparameter prompt roi size $S_{ps}$ in the local prompt module are conducted. This hyperparameter determines the uniform size of the target instances after cropping the corresponding image regions in the original image. If this size is too small it will not be able to fully exploit the texture detail information of the target instances in the original image, and too large it brings redundant computational consumption, so we set up an ablation experiment of this hyperparameter to determine an appropriate value. From the change of the curves in \autoref{fig:ablation_img_roi_size}, this size before 28, with the increase of the size, the accuracy of both detection $AP^{d}$ and segmentation $AP^{s}$ continue to increase, after more than 28, the two accuracy metrics do not have a significant change, so we determine the value of the hyperparameter $S_{ps}$ as 28.

\subsubsection{Space and time complexity analysis}

The space and time complexity comparisons of our proposed approach are shown in the last two columns of \autoref{tab:abla}. We use the number of model parameters as model size to illustrate model space complexity, as shown in the last column of \autoref{tab:abla}. We adopt inference speed $FPS$ to illustrate the time complexity of our proposed approach, as shown in the second-last column of \autoref{tab:abla}.
Although our proposed components bring a little parameter and inference time improvement, it is still able to satisfy the needs of instance segmentation applications, and the improvement of instance segmentation performance is more prominent, therefore, our proposed method is worthwhile.

In conclusion, each component of our proposed prompt paradigm is effective for remote sensing instance segmentation.

\subsection{Qualitative comparison}
\begin{figure}[t]
\centering
\includegraphics[width=3.4in]{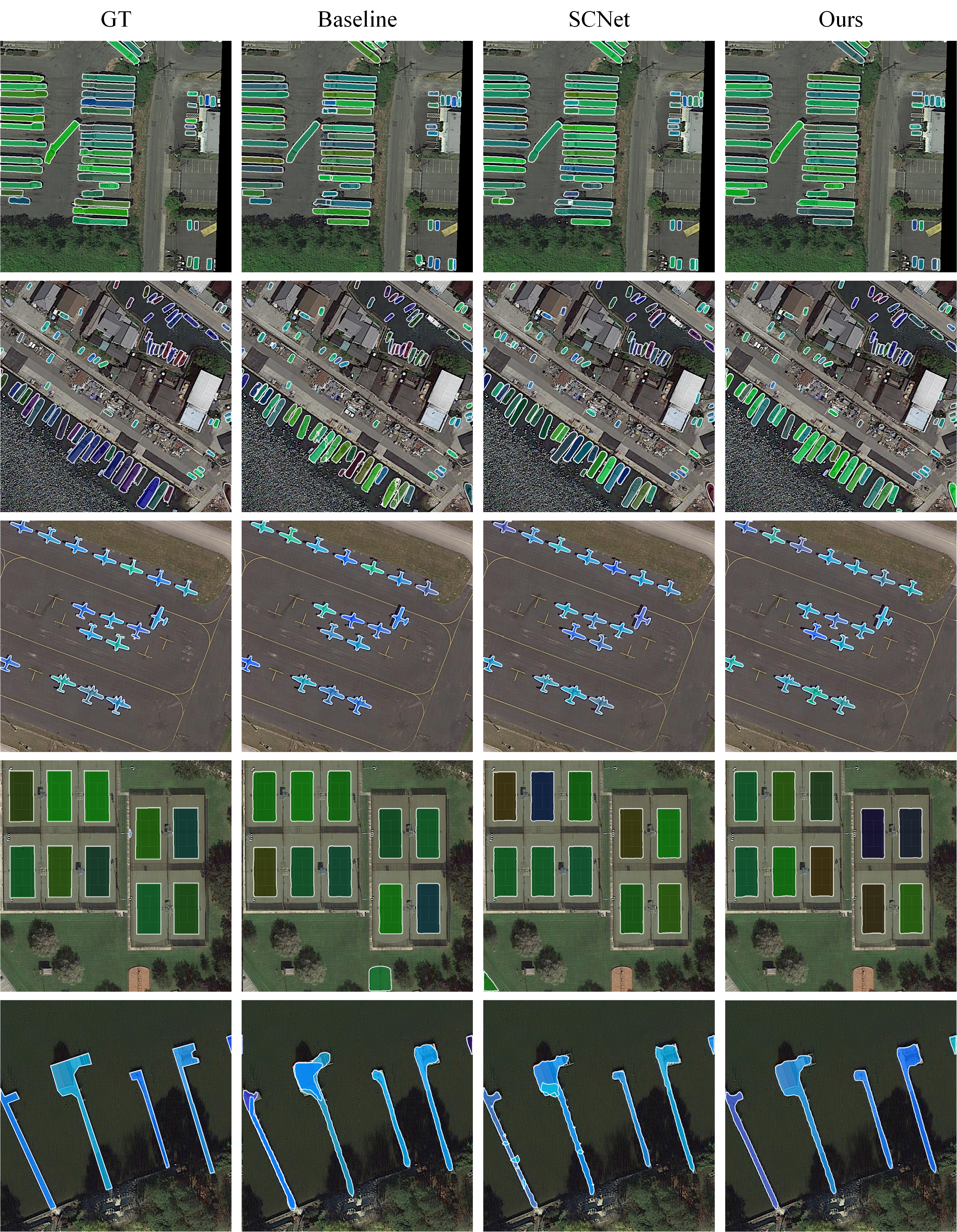}
\caption{The visualisation results of optical image instance segmentation on ISAID dataset. }
\label{fig:opt_vis}
\end{figure}
For instance segmentation of optical remote sensing images, the comparison of visualisation is shown in \autoref{fig:opt_vis}. What can be observed from the comparisons is that our proposed approach has several significant improvements over the baseline model Mask-RCNN and the existing SCNet, including better mask integrity, fewer false alarms, and more accurate mask boundaries.

\begin{figure}[t]
\centering
\includegraphics[width=3.4in]{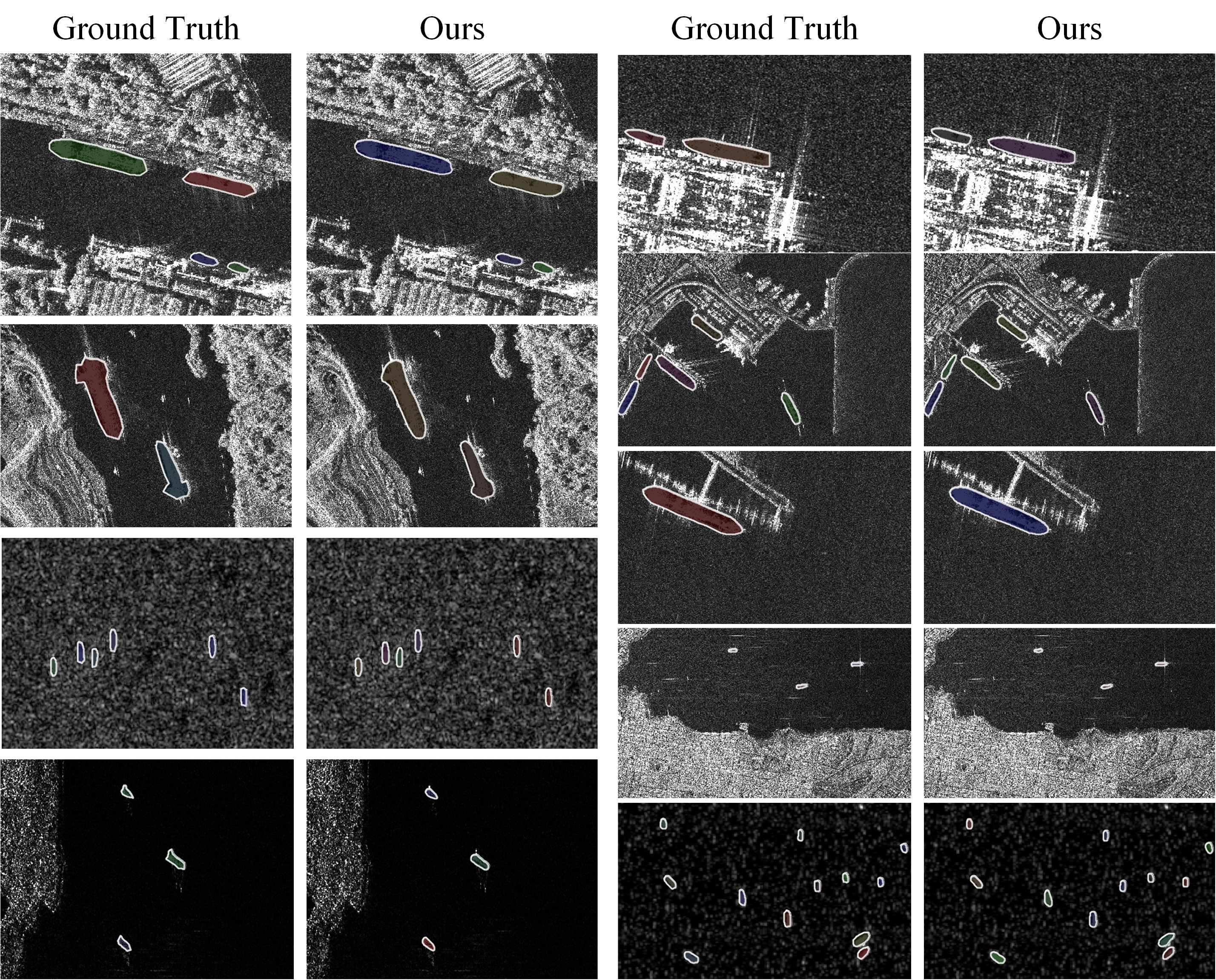}
\caption{The visualisation results of SAR image instance segmentation on SSDD dataset. }
\label{fig:sar_vis}
\end{figure}
For instance segmentation of SAR images, the visualisation results are shown in \autoref{fig:sar_vis}.
From the visualisation, it can be observed that for the ship samples in the SAR images, our proposed approach performs well for a wide range of scenarios such as different scales and complex backgrounds, including near-shore or far-shore. Moreover, the segmentation boundary is accurate compared to the ground truth.

From the above qualitative results, it is clear that our proposed prompt paradigm is a competitive approach for instance segmentation of remote sensing images, both optical and SAR images.

\subsection{Quantitative Comparison}
 In this section, we conduct a quantitative comparison with other state-of-the-art works on four benchmark datasets: ISAID, NWPU VHR-10, SSDD and HRSID to illustrate that our proposed approach is a competitive instance segmentation model for remote sensing images.

\subsubsection{Instance Segmentation Performance}
\
\newline
\indent\textbf{ISAID:}
\autoref{tab:isaid_mask} lists the instance segmentation comparison with other state-of-the-art works on ISAID dataset. We conduct experiments by embedding our proposed prompt paradigm into three models: mask rcnn\cite{he2017mask}, cascade mask rcnn\cite{cai2018cascade}, and scnet\cite{vu2021scnet}. Experimental results based on these three models show that our proposed approach all lead to an improvement in instance segmentation performance, by 2.7\% on Mask-RCNN, 2.3\% on cascade mask rcnn, and 0.7\% on scnet. The best performance scnet embedded with our approach outperforms other state-of-the-art works in $AP$, $AP_{50}$, $AP_{75}$, $AP_{s}$ metrics. And the cascade mask rcnn embedded with our approach outperforms other state-of-the-art works in $AP_m$ evaluation metrics. From the above comparative analysis, it can be concluded that our proposed approach is a competitive remote sensing image instance segmentation work on the iSAID dataset.

\textbf{NWPU VHR-10:}
The results of the comparison with other state-of-the-art works on the NWPU VHR-10 dataset are presented in \autoref{tab:nwpu_mask}. According to the comparative results in the table, the cascade mask rcnn\cite{cai2018cascade} embedded with our proposed approach significantly outperforms the other works, both in terms of the evaluation metrics $AP$ reflected in the performance on the overall data as well as the $AP_s$ reflected in the performance on the small instance set, the $AP_m$ on the medium instance set, and the $AP_l$ on the large instance set. It can be concluded that our proposed prompt paradigm is a new state-of-the-art instance segmentation work on the NWPU VHR-10 dataset.

\textbf{SSDD:}
\autoref{tab:ssdd_mask} lists the comparison of our proposed approach with other state-of-the-art works on the SSDD dataset. Compared to other state-of-the-art works, cascade mask rcnn embedded in our proposed prompt paradigm outperforms the performance evaluation metric $AP$ on the overall dataset and $AP_m$ on the medium instance set. For the evaluation metric $AP_s$ on the small instance set, although slightly lower than the last state-of-the-art work RSIISN\cite{ye2023remote}, it is significantly higher than the other works. It can be concluded that our proposed prompt paradigm is a competitive instance segmentation work for remote sensing SAR images on the SSDD dataset.

\textbf{HRSID:}
\autoref{tab:hrsid_mask} lists the comparison of our proposed approach with other state-of-the-art works on the HRSID dataset. From the comparison results, it can be observed that the cascade mask rcnn embedded in our proposed prompt paradigm significantly outperforms other state-of-the-art works. On the overall dataset, the performance improvement over the last state-of-the-art work RSIISN\cite{ye2023remote}, is $AP$: 3.5\%; on the small instance set, the performance improvement is $AP_s$: 4.0\%. While slightly lower than RSIISN on the medium and large target sets, it is still significantly higher than the other work. Taken together, our proposed prompt paradigm is an advanced remote sensing SAR image instance segmentation work on the HRSID dataset.

\subsubsection{Object Detection Performance}
\
\newline
\indent The facilitation between different tasks plays a crucial role in optimising the learning of the model, e.g., the classification task provides instance category information that facilitates the accuracy of instance segmentation, while instance segmentation facilitates object detection in spatial localisation. Therefore, we establish a study on the performance of our proposed approach for the object detection task. After examining the comparison results listed in \autoref{tab:isaid_bbox}, \autoref{tab:nwpu_bbox}, \autoref{tab:ssdd_bbox}, and \autoref{tab:hrsid_bbox}, what can be found is that our proposed approach outperforms the previous works in object detection on all three datasets except the SSDD dataset. From the comparison results in \autoref{tab:ssdd_bbox}, although our approach is not optimal in terms of $AP$ evaluation metrics, it is superior to other works in terms of $AP_s$ for small object sets and $AP_l$ metrics for medium object sets. The above analyses show that our proposed approach has a positive impact on object detection task in remote sensing images. The synergistic facilitation between these interrelated tasks helps to improve the model to address complex visual problems in remote sensing images.

To sum up, our proposed prompt paradigm is a competitive approach for instance segmentation of remote sensing images, both in optical remote sensing data, and SAR remote sensing data.
\begin{table}
\footnotesize
\renewcommand{\arraystretch}{1.5}
\centering
\caption{Quantitative instance segmentation performance (mask $AP$) on ISAID dataset. 
$^{\#}$ means the backbone is ResNet101.}
\label{tab:isaid_mask}
\resizebox{\linewidth}{!}{
\begin{tabular}{ccccccc}
\cline{1-7}
 Model& $AP$ & $AP_{50}$ & $AP_{75}$ & $AP_{s}$ & $AP_{m}$ & $AP_{l}$\\                              
\cline{1-7} 
YOLACT\cite{bolya2019yolact}                       & 22.3 & 43.3 & 19.1 & 8.4  & 31.8 & 40.4 \\
Box2Mask-C$^{\#}$\cite{li2022box2mask}            & 26.6 & 50.6 & 23.8 & 10.6 & 33.6 & 47.4 \\
Ref.\cite{luo2022elliptic}                         & 29.4 & 54.5 & 27.8 & 15.5 & 37.8 & 42.0 \\
Mask-RCNN\cite{he2017mask}   & 33.8 & 56.6 & 35.7 & 19.4 & 40.1 & 45.6 \\ 
\scriptsize{Mask Scoring RCNN}\cite{huang2019mask}            & 35.9 & 57.7 & 38.4 & 20.8 & 44.3 & 51.5   \\      
PointRend \cite{kirillov2020pointrend}                    & 35.6 & 59.0 & 37.3 & 20.3 & 44.8 & 52.9   \\ 
\scriptsize{Cascade Mask RCNN}\cite{cai2018cascade}            & 35.6 & 57.8 & 38.0 & 20.8 & 44.3 & 52.7   \\ 
HTC\cite{chen2019hybrid}                          & 37.4 & 60.2 & 40.1 & 23.5 & 44.6 & 53.5     \\
SCNet\cite{vu2021scnet}                        & 37.3 & 59.5 & 40.3 & 23.3 & 44.8 & 52.3   \\
RSIISN\cite{ye2023remote}                       & 37.6 & 60.8 & 41.4 & 22.4 & \textbf{47.7} & 53.4  \\ \cline{1-7}
Mask RCNN(ours)              & 36.5           & 60.4   & 39.9  & 21.1 & 44.1 & 52.2    \\
\scriptsize{Cascade Mask RCNN(ours)}       &  37.9     & 61.3  & 40.3 & 22.0& 44.0 & \textbf{54.9}    \\
SCNet(ours)                               & \textbf{38.0}  & \textbf{63.3} & \textbf{41.5} & \textbf{23.8} & 45.7 & 54.3    \\
\cline{1-7}
\end{tabular}
}
\end{table}

\begin{table}
\footnotesize
\renewcommand{\arraystretch}{1.5}
\tabcolsep=0.30cm
\centering
\caption{Quantitative object detection performance (bbox $AP$) on ISAID dataset. '-' means that the value is not reported in the paper. }
\label{tab:isaid_bbox}
\begin{tabular}{ccccc}
\cline{1-5}
 Model& $AP$ & $AP_{s}$ & $AP_{m}$ & $AP_{l}$\\                              
\cline{1-5} 
YOLACT\cite{bolya2019yolact}                        & 28.8 & 16.7 & 33.7 & 39.7  \\
Mask-RCNN\cite{he2017mask}                          & 39.6 & 25.1 & 46.4 & 47.7 \\ 
PointRend\cite{kirillov2020pointrend}               & 40.2 & 26.2 & 48.0 & 52.1   \\ 
Mask Scoring RCNN\cite{huang2019mask}               & 40.5 & 25.8 & 48.8 & 53.3   \\      
Ref.\cite{luo2022elliptic}                          & -    & 27.4 & 47.1 & 31.1   \\ 
Cascade Mask RCNN\cite{cai2018cascade}              & 43.1 & 28.1 & 48.7 & 56.0   \\
RSIISN\cite{ye2023remote}                           & 46.6 & 32.3 & 54.8 & 61.6 \\
Mask-RCNN(ours)                                     & 42.8 & 27.5 & 49.4 & 56.4 \\
SCNet(ours)                                         & 44.4 & \textbf{29.7} & \textbf{51.0} & 56.6 \\
Cascade Mask RCNN(ours)                             & \textbf{44.8} & 29.1 & 49.7 & \textbf{60.0}    \\
\cline{1-5}
\end{tabular}
\end{table}

\begin{table}
\footnotesize
\renewcommand{\arraystretch}{1.5}
\tabcolsep=0.30cm
\centering
\caption{Quantitative instance segmentation performance (mask $AP$) on the test set of NWPU VHR-10 dataset. '-' means that the value is not reported in the paper. }
\label{tab:nwpu_mask}
\begin{tabular}{ccccc}
\cline{1-5}
 Model& $AP$ & $AP_{s}$ & $AP_{m}$ & $AP_{l}$\\                              
\cline{1-5} 
YOLACT\cite{bolya2019yolact}                        & 45.2 & 18.9 & 44.0 & 63.1  \\
Mask-RCNN\cite{he2017mask}                          & 54.9 & 33.7 & 55.1 & 52.3 \\ 
PointRend\cite{kirillov2020pointrend}               & 59.9 & 46.3 & 60.5 & 54.0   \\ 
Mask Scoring RCNN\cite{huang2019mask}               & 57.0 & 46.7 & 56.9 & 55.8   \\      
Cascade Mask RCNN\cite{cai2018cascade}              & 58.1 & 44.2 & 58.4 & 47.1   \\
HRNetV2p-W32\cite{kumar2022accurate}                & 65.1 & 49.5 & 54.7 & 69.8   \\
FB-ISNet\cite{su2022faster}                         & 66.5 & - & - & - \\
Cascade Mask RCNN(ours)                             & \textbf{66.9} & \textbf{53.6} & \textbf{65.9} & \textbf{77.3}    \\
\cline{1-5}
\end{tabular}
\end{table}

\begin{table}
\footnotesize
\renewcommand{\arraystretch}{1.5}
\tabcolsep=0.30cm
\centering
\caption{Quantitative object detection performance (bbox $AP$) on the test set of NWPU VHR-10 dataset. '-' means that the value is not reported in the paper. }
\label{tab:nwpu_bbox}
\begin{tabular}{ccccc}
\cline{1-5}
 Model& $AP$ & $AP_{s}$ & $AP_{m}$ & $AP_{l}$\\                              
\cline{1-5} 
YOLACT\cite{bolya2019yolact}                        & 52.7 & 51.8 & 52.4 & 50.8  \\
Mask-RCNN\cite{he2017mask}                          & 62.7 & 66.3 & 64.4 & 53.6 \\ 
PointRend\cite{kirillov2020pointrend}               & 59.7 & 61.7 & 60.6 & 45.2   \\ 
Mask Scoring RCNN\cite{huang2019mask}               & 61.9 & 57.9 & 62.6 & 48.7   \\      
Ref.\cite{luo2022elliptic}                          & 53.2    & - & - & -   \\ 
Cascade Mask RCNN\cite{cai2018cascade}              & 65.7 & 56.6 & 66.7 & 52.4   \\
WSODet\cite{tan2023wsodet}                          & 61.3 & - & - & - \\
DAL   \cite{zhang2023movable}                       & 63.93 & - & - & - \\
RSIISN\cite{ye2023remote}                           & 67.6  & 64.9 & 68.4 & 55.9 \\
Cascade Mask RCNN(ours)                             & \textbf{71.3} & \textbf{71.1} & \textbf{71.6} & \textbf{67.7}    \\
\cline{1-5}
\end{tabular}
\end{table}

\begin{table}
\footnotesize
\renewcommand{\arraystretch}{1.5}
\tabcolsep=0.30cm
\centering
\caption{Quantitative instance segmentation performance (mask $AP$) on the test set of SSDD dataset. '-' means that the value is not reported in the paper. }
\label{tab:ssdd_mask}
\begin{tabular}{ccccc}
\cline{1-5}
 Model& $AP$ & $AP_{s}$ & $AP_{m}$ & $AP_{l}$\\                              
\cline{1-5} 
YOLACT\cite{bolya2019yolact}                        & 56.3 & 57.9 & 53.6 & 17.6  \\
FL-CSE-ROIE\cite{zhang2022full}                     & 62.6 & 63.3 & 61.2 & 75.0 \\
Mask-RCNN\cite{he2017mask}                          & 64.2 & 66.2 & 58.3 & 27.6 \\ 
MAI-SE-Net\cite{zhang2022mask}                      & 63.0 & 63.3 & 62.5 & 47.7 \\
PointRend\cite{kirillov2020pointrend}               & 65.4 & 67.4 & 59.8 & 39.5   \\ 
Mask Scoring RCNN\cite{huang2019mask}               & 64.6 & 66.9 & 57.6 & 15.0   \\      
Cascade Mask RCNN\cite{cai2018cascade}              & 64.9 & 67.4 & 57.5 & 38.7   \\
C-SE Mask RCNN\cite{zhang2022contextual}            & 58.6 & 58.3 & 60.7 & 26.7 \\
EMIN\cite{zhang2022enhanced}                        & 61.7 & 62.1 & 61.3 & 40.1 \\
RSIISN\cite{ye2023remote}                           & 68.4 & \textbf{69.8} & 63.7 & \textbf{41.0} \\
Cascade Mask RCNN(ours)                             & \textbf{70.0} & 69.1 & \textbf{73.4} & 33.6    \\
\cline{1-5}
\end{tabular}
\end{table}

\begin{table}
\footnotesize
\renewcommand{\arraystretch}{1.5}
\tabcolsep=0.30cm
\centering
\caption{Quantitative object detection performance (bbox $AP$) on the test set of SSDD dataset. '-' means that the value is not reported in the paper. }
\label{tab:ssdd_bbox}
\begin{tabular}{ccccc}
\cline{1-5}
 Model& $AP$ & $AP_{s}$ & $AP_{m}$ & $AP_{l}$\\                              
\cline{1-5} 
YOLACT\cite{bolya2019yolact}                        & 60.4 & 64.5 & 49.8 & 23.7  \\
Mask-RCNN\cite{he2017mask}                          & 68.6 & 70.9 & 61.6 & 8.9 \\ 
PointRend\cite{kirillov2020pointrend}               & 67.0 & 70.1 & 57.2 & 13.9   \\ 
Mask Scoring RCNN\cite{huang2019mask}               & \textbf{84.5} & 71.0 & 61.5 & 7.9   \\      
Ref.\cite{luo2022elliptic}                          & 66.4 & 64.3 & 70.2 & \textbf{72.0}   \\ 
Cascade Mask RCNN\cite{cai2018cascade}              & 69.1 & 71.4 & 62.8 & 9.2   \\
RSIISN\cite{ye2023remote}                           & 70.3 & 72.7 & 62.9 & 33.9  \\
Cascade Mask RCNN(ours)                             & 73.6 & \textbf{72.8} & \textbf{77.0} & 42.7    \\
\cline{1-5}
\end{tabular}
\end{table}

\begin{table}
\footnotesize
\renewcommand{\arraystretch}{1.5}
\tabcolsep=0.30cm
\centering
\caption{Quantitative instance segmentation performance (mask $AP$) on HRSID dataset. '-' means that the value is not reported in the paper. }
\label{tab:hrsid_mask}
\begin{tabular}{ccccc}
\cline{1-5}
 Model& $AP$ & $AP_{s}$ & $AP_{m}$ & $AP_{l}$\\                              
\cline{1-5} 
YOLACT\cite{bolya2019yolact}                        & 31.8 & 30.0 & 50.1 & 13.6  \\
Mask-RCNN\cite{he2017mask}                          & 53.4 & 52.9 & 62.2 & 15.6 \\ 
PointRend\cite{kirillov2020pointrend}               & 51.8 & 51.1 & 61.5 & 19.6   \\ 
Mask Scoring RCNN\cite{huang2019mask}               & 53.5 & 53.1 & 61.6 & 18.1   \\      
Cascade Mask RCNN\cite{cai2018cascade}              & 54.2 & 53.7 & 62.4 & 22.2   \\
RSIISN\cite{ye2023remote}                           & 55.8 & 54.8 & \textbf{66.7} & \textbf{35.2} \\
Cascade Mask RCNN(ours)                             & \textbf{59.3} & \textbf{58.8} & 66.2 & 33.7    \\
\cline{1-5}
\end{tabular}
\end{table}

\begin{table}
\footnotesize
\renewcommand{\arraystretch}{1.5}
\tabcolsep=0.30cm
\centering
\caption{Quantitative object detection performance (bbox $AP$) on HRSID dataset. '-' means that the value is not reported in the paper. }
\label{tab:hrsid_bbox}
\begin{tabular}{ccccc}
\cline{1-5}
 Model& $AP$ & $AP_{s}$ & $AP_{m}$ & $AP_{l}$\\                              
\cline{1-5} 
YOLACT\cite{bolya2019yolact}                        & 53.2 & 54.8 & 53.5 & 21.2  \\
Mask-RCNN\cite{he2017mask}                          & 60.9 & 62.0 & 64.7 & 12.9 \\ 
PointRend\cite{kirillov2020pointrend}               & 57.7 & 58.8 & 60.2 & 11.7   \\ 
Mask Scoring RCNN\cite{huang2019mask}               & 60.1 & 61.3 & 64.5 & 12.8   \\      
Ref.\cite{luo2022elliptic}                          & 55.2 & 38.9 & 70.2 & 65.0   \\ 
Cascade Mask RCNN\cite{cai2018cascade}              & 64.4 & 65.3 & 66.7 & 21.6   \\
RSIISN\cite{ye2023remote}                           & 63.9 & 64.4 & 68.4 & 36.9  \\
Cascade Mask RCNN(ours)                             & \textbf{70.4} & \textbf{71.5} & \textbf{71.4} & \textbf{43.4}   \\
\cline{1-5}
\end{tabular}
\end{table}

\subsection{Instance-specific boxes prompt for promptable instance segmentation}

The above experimental chapters demonstrate that our proposed approach is beneficial for the remote sensing instance segmentation task. It is worth mentioning that our proposed approach supports not only automatic instance segmentation, but also promptable instance segmentation with boxes prompt. Our proposed prompt paradigm enables existing instance segmentation models to be extended to support the promptable instance segmentation task. This section evaluates the performance of our proposed promptable instance segmentation from both qualitative and quantitative perspectives. 

\begin{figure}[t]
\centering
\includegraphics[width=3.4in]{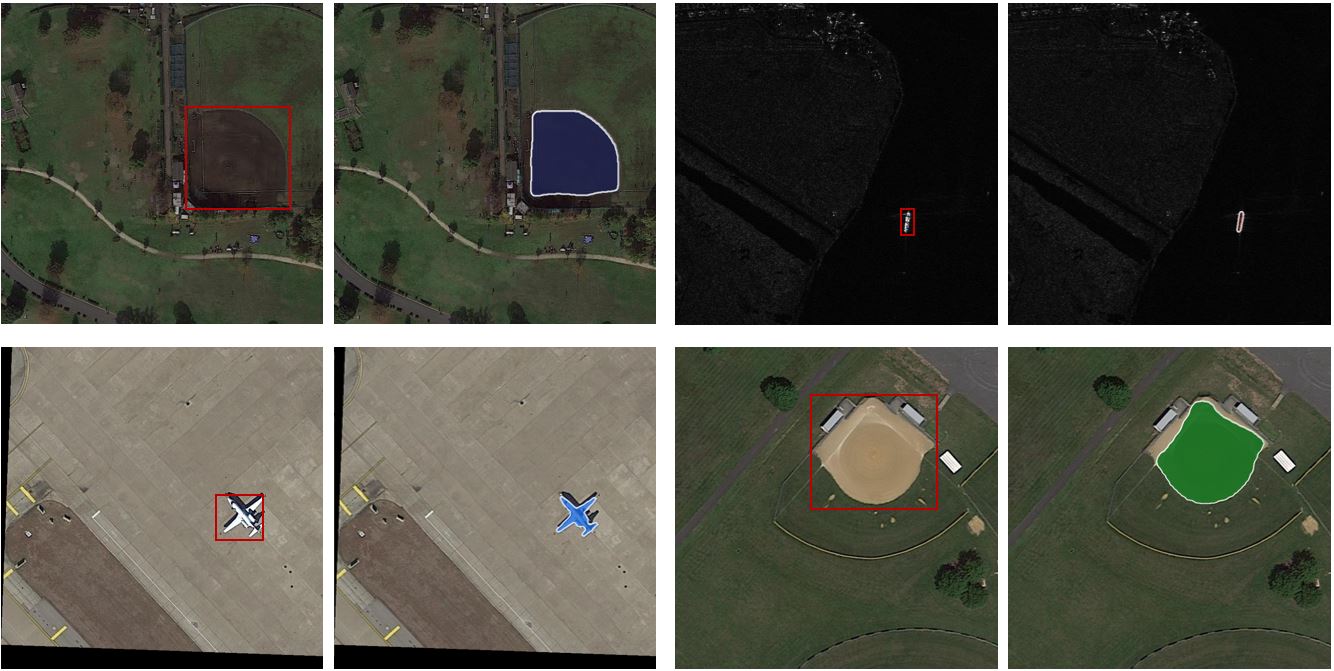}
\caption{Examples of promptable instance segmentation. On the left is the given boxes prompt and on the right is the corresponding mask results. }
\label{fig:promptable_is}
\end{figure}

The qualitative promptable instance segmentation with the boxes prompt is shown in \autoref{fig:promptable_is}. Given the boxes prompt, our promptable instance segmentation model can complete the corresponding instance segmentation. Firstly the model completes the feature extraction of the image encoder based on the input image, and then, based on the input boxes prompt, the prompt encoder together with the instance segmentation decoder completes the instance segmentation of the corresponding regions. It is worth mentioning that after completing the extraction of the encoder, the promptable instance segmentation in the boxes prompt takes only 40 ms. 

The quantitative promptable instance segmentation evaluation results are shown in \autoref{tab:promptable_mask}. For some missed instances of automatic instance segmentation, we input the boxes prompt into the promptable instance segmentation model, output the mask, and finally compute the evaluation metrics together with the automatic instance segmentation results. The reason for the few improvements is that the number of missed instances in the automatic instance segmentation model with our prompt paradigm is less compared to the number of recalled instances, but it is enough to prove that the promptable instance segmentation model constructed on the basis of our proposed prompt paradigm is effective. 
\begin{table}
\footnotesize
\renewcommand{\arraystretch}{1.5}
\tabcolsep=0.40cm
\centering
\caption{Quantitative instance segmentation performance (mask $AP$) about promptable instance segmentation. 'IS' indicates instance segmentation.  }
\label{tab:promptable_mask}
\begin{tabular}{cccc}
\cline{1-4}
 Model & Baseline & IS & Promptable IS \\                              
\cline{1-4} 
Mask $AP$  & 58.1  & 66.9  & 67.7   \\
\cline{1-4}
\end{tabular}
\end{table}

The above analyses demonstrate that our proposed prompt paradigm can be extended to promptable instance segmentation models with some effectiveness.

\section{Conclusion}

In this paper, we propose a novel prompt paradigm for the problem of unbalanced foreground and background pixel ratios and limited instance sizes in remote sensing images, as well as the harm of downsampling in the existing deep feature extraction paradigms to address this problem. The local prompt module fully exploits the texture information of the instances in the input image as a local prompt, and the global-to-local prompt module models the contextual information from the global to the local part of the instances as a global prompt, while a constrained loss function on the scale of the proposals is used to better exploit the potential of the two prompt modules. Thorough experiments demonstrate that our proposed approach is effective and competitive for remote sensing image instance segmentation.
In future research, we expect to boost the generalisation and versatility of promptable remote sensing image instance segmentation models by constructing more appropriate training strategies on large-scale datasets, but this is also beyond the scope of this paper.

{\small
\bibliographystyle{unsrt}
\bibliography{egbib}
}

\vfill

\end{document}